\documentclass[10pt,letterpaper]{article}
\usepackage[top=0.85in,left=2.75in,footskip=0.75in]{geometry}

\usepackage{changepage}

\usepackage[utf8x]{inputenc}

\usepackage{textcomp,marvosym}

\usepackage{fixltx2e}

\usepackage{amsmath,amssymb}

\usepackage{cite}

\usepackage{nameref,hyperref}

\usepackage[right]{lineno}

\usepackage{color}

\usepackage{microtype}
\DisableLigatures[f]{encoding = *, family = * }

\usepackage{booktabs}

\raggedright
\usepackage[aboveskip=1pt,labelfont=bf,labelsep=period,justification=raggedright,singlelinecheck=off]{caption}

\makeatletter
\renewcommand{\@biblabel}[1]{\quad#1.}
\makeatother

\date{}

\usepackage{lastpage,fancyhdr,graphicx}
\usepackage{epstopdf}
\fancyheadoffset[L]{2.25in}
\fancyfootoffset[L]{2.25in}
\begin{document}
\vspace*{0.2in}

\begin{flushleft}
{\Large
\textbf\newline{Probabilistic Fluorescence-Based Synapse Detection} 
}
\newline
\\
Anish K. Simhal		\textsuperscript{1*},
Cecilia Aguerrebere		\textsuperscript{1},
Forrest Collman		\textsuperscript{2},
Joshua T. Vogelstein		\textsuperscript{3},
Kristina D. Micheva		\textsuperscript{4},
Richard J. Weinberg		\textsuperscript{5},
Stephen J. Smith		\textsuperscript{2},
Guillermo Sapiro		\textsuperscript{1, 6}
\\
\bigskip
\textbf{1} Electrical and Computer Engineering, Duke University, Durham, North Carolina, USA
\\
\textbf{2} Synapse Biology, Allen Institute for Brain Sciences, Seattle, Washington, USA
\\
\textbf{3} Department of Biomedical Engineering, Johns Hopkins University, Baltimore, Maryland, USA
\\
\textbf{4} Molecular and Cellular Physiology, Stanford University School of Medicine, Stanford, California, USA
\\
\textbf{5} Department of Cell Biology and Physiology, University of North Carolina, Chapel Hill, North Carolina, USA
\\
\textbf{6} Department of Biomedical Engineering, Department of Computer Science, Department of Mathematics, Duke University, Durham, North Carolina, USA
\bigskip

%
%





* Corresponding Author: anish.simhal@duke.edu

\end{flushleft}
\section*{Abstract}

Deeper exploration of the brain's vast synaptic networks will require new tools for high-throughput structural and molecular profiling of the diverse populations of synapses that compose those networks.  Fluorescence microscopy (FM) and electron microscopy (EM) offer complementary advantages and disadvantages for single-synapse analysis. FM combines exquisite molecular discrimination capacities with high speed and low cost, but rigorous discrimination between synaptic and non-synaptic fluorescence signals is challenging. In contrast, EM remains the gold standard for reliable identification of a synapse, but offers limited molecular discrimination and is slow and costly. To develop and test single-synapse image analysis methods, we have used datasets from conjugate array tomography (cAT), which provides voxel-conjugate FM and EM (annotated) images of the same individual synapses. We report a novel unsupervised probabilistic method for detection of synapses from multiplex FM (muxFM) image data, and evaluate this method both by comparison to EM gold standard annotated data and by examining its capacity to reproduce known important features of cortical synapse distributions. The proposed probabilistic model-based synapse detector accepts molecular-morphological synapse models as user queries, and delivers a volumetric map of the probability that each voxel represents part of a synapse. Taking human annotation of  cAT EM data as ground truth, we show that our algorithm detects synapses from muxFM data alone as successfully as human annotators seeing only the muxFM data, and accurately reproduces known architectural features of cortical synapse distributions. This approach opens the door to data-driven discovery of new synapse types and their density. We suggest that our probabilistic synapse detector will also be useful for analysis of standard confocal and super-resolution FM images, where EM cross-validation is not practical.


\section*{Author Summary}
Brain function results from communication between neurons connected by complex synaptic networks. Synapses are themselves highly complex and diverse signaling machines, containing protein products of hundreds of different genes, some in hundreds of copies, arranged in precise lattice at each individual synapse. Synapses are fundamental not only to synaptic network function but also to network development, adaptation, and memory.  In addition, abnormalities of synapse numbers or molecular components are implicated in most mental and neurological disorders. Despite their obvious importance, mammalian synapse populations have so far resisted detailed quantitative study. In human brains and most animal nervous systems, synapses are very small and very densely packed: there are approximately 1 billion synapses per cubic millimeter of human cortex. This volumetric density poses very substantial challenges to proteometric analysis at the critical level of the individual synapse. The present work describes new probabilistic image analysis methods for single-synapse analysis of synapse populations in both animal and human brains, in health and disorder.


\section*{Introduction}

Deeper understanding of the basic mechanisms and pathologies of the brain's synaptic networks will require a much clearer and more quantitative understanding of the vast populations of highly diverse individual synapses that compose those networks. As a	step toward such understanding, we here introduce and characterize novel image analysis methods for the high-throughput single-synapse molecular profiling of synapse populations from multiplex fluorescence microscopy (muxFM); in the present report we focus on immunofluorescence array tomography (IF-AT) image data.   Coupled with high-throughput and high-resolution muxFM image acquisition methods, these image analysis approaches will permit single-synapse analysis at scale of billions of synapses, as required to fully characterize normal and pathological specimens of mammalian neocortex, e.g., \cite{Fitzsimmons} \cite{Lichtman}.
\newline
\newline
Single-synapse profiling of large and diverse synapse populations poses  formidable challenges.  Electron microscopy (EM) of  appropriately labeled specimens defines the current `gold standard' for synapse detection: the nanometer	resolution of EM is necessary for the unambiguous identification of defining synaptic features such as presynaptic vesicles, synaptic clefts, and postsynaptic densities. Unfortunately EM data acquisition is technically difficult, slow, burdened by large data processing and storage requirements, and offers only limited capacities to discriminate amongst the hundreds of different synaptic proteins that constitute the synaptic proteome.  In contrast, fluorescence microscopy (FM) of tagged specimens is much faster and less expensive, easier to segment for analysis, and offers much greater molecular discrimination power. Unfortunately, the ability of FM to detect and discriminate individual synapses is compromised by resolution limits and the close crowding of synapses in most neural tissue specimens of interest.  Robust FM detection of synapses is nonetheless potentially possible by combining	 measures that extend resolution limits and multiplexing for localization and co-localization of synaptic markers.
\newline
\newline
In designing the algorithm and software reported here, we first relied on images acquired with conjugate array tomography (cAT), which combines the strengths of	 FM-AT with those of electron microscopic array tomography (EM-AT), allowing both EM and muxFM imaging of individual cortical specimens.  Array tomography's ultrathin physical sectioning provides z-axis resolution far beyond the light microscopic diffraction limit, as well as high sensitivity and high lateral resolution, while greatly simplifying voxel-conjugate registration of FM-AT and EM-AT images.  FM-AT can	 moreover multiplex large numbers of synaptic markers by its combination of sequential and spectral label multiplexing.  Thus, cAT provides an ideal platform   for the development and rigorous design and testing of algorithms aimed at single-synapse molecular analysis and population molecular profiling.
\newline
\newline
The remarkable structural and molecular diversity within mammalian synapse populations challenges our present biological understanding of how to define a synapse.  Difficulties also arise from a very broad	 distribution of synapse size, with the smallest synapses occurring at the highest frequencies.  Thus, detection of a synapse inevitably involves setting some minimum-size criterion for any candidate cell-cell contact specialization to qualify as a synapse.  For FM, the `size' metric is typically the intensity of one or more fluorescent synaptic protein tags. The fact that there are clearly non-synaptic `backgrounds,' and that the observed size distributions are log-normal, enforces high sensitivity of synapse detection on some rather arbitrary threshold minimum size value.  This sensitivity in turn makes key results of widespread interest, such as the synaptic density in a region or the presence/absence of a synapse at a given microscopic site, uncomfortably dependent on that same size threshold value.  The probabilistic synapse detector proposed here may lead both to relief from such arbitrary-threshold (parameters) and to improvements in our biological understanding of what defines a synapse.
\newline
\newline
The unsupervised probabilistic synapse detector reported here accepts molecular-morphological synapse models in the form of user queries, and   delivers a	volumetric map of the probability that each voxel represents part of a synapse. These maps can then be used directly to detect, classify, and map putative synapses, with confidence statistics for each.  Taking human annotation of cAT EM data as ground truth, we show that our algorithm detects synapses from muxFM data alone as effectively as human annotators (while seeing only the muxFM data), and can reproduce known architectural features of cortical synapse distributions. The algorithm is actually validated with the most comprehensive AT  datasets currently available.  Though we  here address only array tomography image data, our probabilistic synapse detector may also be useful for analysis of widely available confocal and super-resolution FM images.

\section*{Methods}

\subsection*{\textit{Overview}}
\begin{figure}
\centering
\includegraphics[width=1\textwidth]{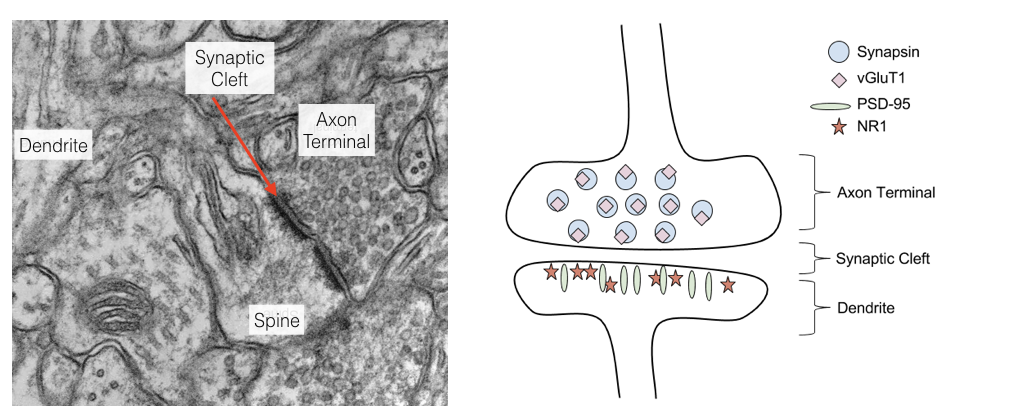}
\caption{\textbf{Excitatory Synapse - }  {\it Left:}  In this high magnification image, the synaptic cleft is clearly visible, and close examination reveals the lipid bilayer of the plasma membranes. The (perforated) excitatory axospinous synapse is on the left of the figure. A spine apparatus is visible in the body of the spine. A thick spine neck is visible joining with its parent dendrite, on the bottom right. The dendrite has endoplasmic reticulum in it. No other obvious synapses are visible, but just to the right and above the spine is a profile (maybe an axon) that has microtubules cut almost longitudinally.  {\it Right:} Cartoon diagramming the molecular architecture of a excitatory PSD-95-expressing synapse  \cite{Weiler}.  Basic biological knowledge about synapse structure and protein composition as depicted in this figure is used to inform the proposed query-based probabilistic algorithm.}
\label{fig:Chemical_Synapse}
\end{figure}

\begin{figure}
\centering
\includegraphics[width=1\textwidth]{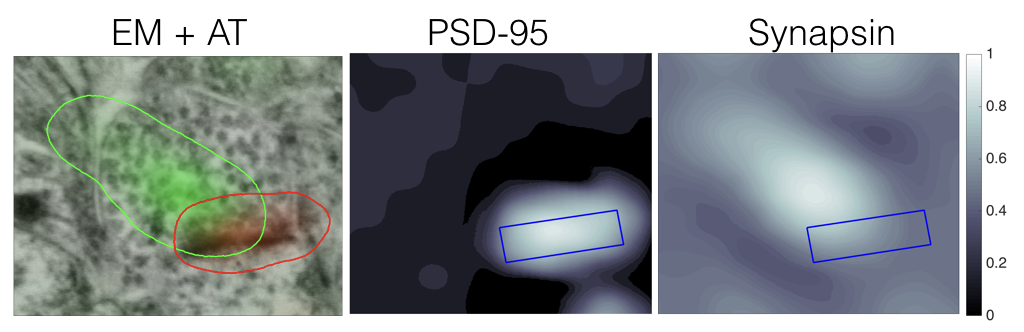}
\caption{\textbf{cAT Data} - Three $1.277$ $\mu$ m $\times 1.186$ $\mu$ m images showing conjugate EM and IF data. The left panel has synapsin (green) and PSD-95 (red) data overlaid, marked by the colored boundary lines. The presence of both presynaptic and postsynaptic channels indicates the presence of a synapse with high probability. The center panel shows the PSD-95 IF image and the right panel shows the synapsin IF image. On both images, the EM-identified synaptic cleft is marked by a blue box.
}
\label{fig:synapseOverview}
\end{figure}

\begin{figure}
\centering
\includegraphics[width=1\textwidth]{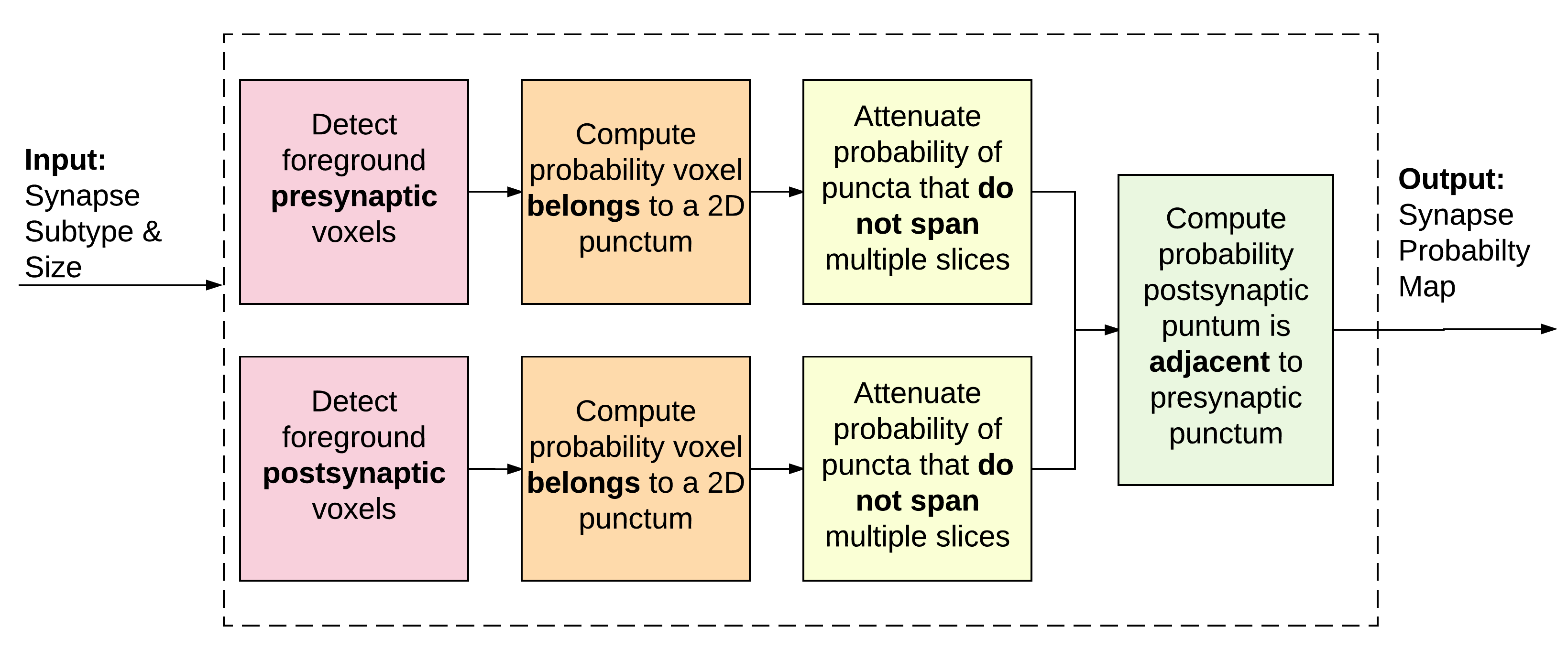}
\caption{\textbf{Proposed Method Flowchart} - Fundamental steps of the proposed probabilistic synapse detection algorithm.}
\label{fig:pipeline}
\end{figure}

The proposed algorithm is inspired from biological knowledge of synapse characteristics. Notwithstanding immense complexity and limited understanding, strong evidence suggests two major synaptic classes: excitatory and inhibitory. Synapses include two major structural components: a presynaptic terminal and a postsynaptic dendrite. Detecting synapses using data from immunofluorescence imaging involves identifying such adjacent presynaptic and postsynaptic antibody markers, as shown in Fig \ref{fig:Chemical_Synapse}, which diagrams the locations of four major excitatory synaptic proteins.  Fig \ref{fig:synapseOverview} is an example of an excitatory synapse with images of presynaptic and postsynaptic antibody markers overlaid upon an EM image.  
\newline 
\newline
Manual synapse identification involves determining the punctum size and brightness in one channel, and then considering adjacency to similarly-defined puncta in other channels.  However, without corresponding EM data, detections using only IF data have an associated degree of uncertainty.  Thus, we propose a query-based probabilistic synapse detection method that reflects the thought process underlying expert manual synapse detection.
\newline
\newline
The first step is to distinguish signal from background noise. This calculation encodes the probability that the pixel value represents authentic antigen detection.  This indicates the probability that the antibody is detecting its corresponding protein antigen.  The second step is to determine whether the foreground pixels correspond to a 2D punctum, since photons emanating only from a single pixel  usually reflect noise. Therefore, adjacent `positive' pixels, more likely to reflect a synaptic punctum, are augmented.  Third, puncta that span  multiple slices have a higher probability of belonging to a synapse than those that do not. To visualize this effect, the probability of a punctum belonging to a synapse is attenuated based on whether the prospective punctum spans multiple slices. The last step in computing the synapse probability map is to evaluate the presence of adjacent presynaptic and postsynaptic puncta by correlating the corresponding IF volumes.  This produces a probability map, where the value at each voxel is the probability it belongs to a synapse.  This algorithm provides a general framework for the evaluation of a wide variety of synapse subtypes, user-defined by setting the presynaptic and postsynaptic antibodies and puncta size.
\newline
\newline
The following sections describe in detail each step in the process, as diagrammed in Fig \ref{fig:pipeline}.

\subsection*{\textit{Step 1: Computation of Foreground Probability}}
\begin{figure}
\centering
\includegraphics[width=1\textwidth]{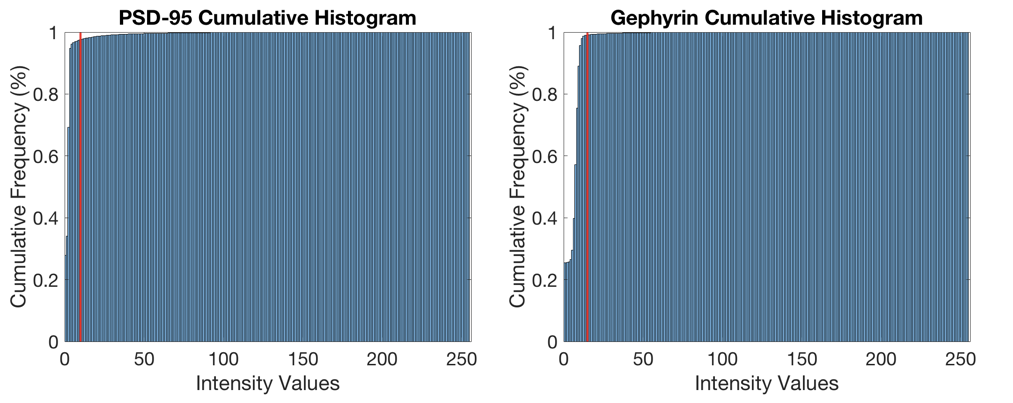}
\caption{\textbf{Raw IF Data Histograms} - These cumulative histograms show that typically $\sim98$\% of voxels in the dataset lie below the threshold line indicated in red.  The threshold lines are estimates based on visual inspection of the data.}
\label{fig:thresholdHistograms}
\end{figure}

\begin{figure}
\centering
\includegraphics[width=1\textwidth]{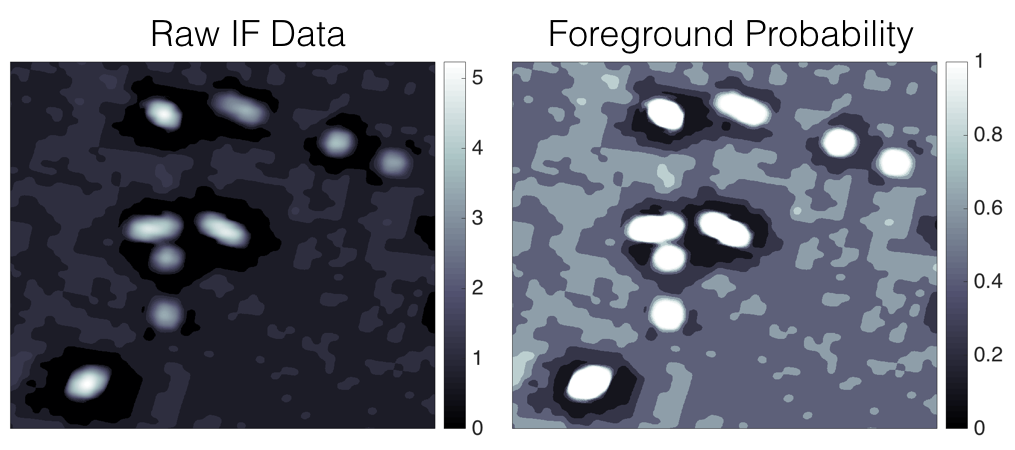}
\caption{\textbf{Foreground Probability} - Cutout of size $2261 \times 2501$ pixels or $5.268 \times 5.827 \mu$ m of the logarithm (for visualization purposes) of the IF raw data (left) and the corresponding image of the foreground probability map (right) of one slice of the PSD-95 antibody. The proposed approach clearly differentiates the high probability bright pixels from the background low probability pixels.  The `dark' rings around the puncta are an artifact of the deconvolution performed prior to image alignment, and its spatial extend has been taken into account in the spatially-oriented next steps of the algorithm.  The AT data appears `quantized'  because it has been upsampled from its native 100 nm per pixel resolution to 2.33 nm per pixel to align the AT data with the EM data. }
\label{fig:raw_prob} 
\end{figure}

Raw immunofluorescence image data is noisy; for example, speckles of the antibody markers often bind with cellular structures not associated with synapses, such as mitochondria.  In addition, fluorescence imagery contains signal from sources other than fluorescently-labelled antibodies, e.g. from background autofluorescence.  Finally, all digital imagery contains inherent noise from sources such as camera read noise and photon shot noise. The noise produced by these sources is usually smaller in magnitude than that originating from authentic synaptic labeling, but it cannot simply be filtered out and dismissed from consideration, since the signal may originate from a true synaptic site, and we want to allow for the possibility that a concordance of weak evidence will lead to the detection of a synapse.  Thus, the first step of the algorithm consists of differentiating the bright voxels, the foreground (potential objects of interest), from a noisy background in a probabilistic fashion. 
\newline
\newline
IF data volumes, when stained for synaptic markers, are also extremely sparse - approximately 2\% of the voxels in the dataset belong to the foreground, as indicated in Fig \ref{fig:thresholdHistograms}.   Therefore, the IF image volume can be used to approximate the distribution of the background noise. 
\newline 
\newline
Let $v(x,y,z)$ be the intensity value of a voxel at position $(x,y)$ in slice $z$, for a given channel of the IF data.  A probabilistic model, $p_{B}$, is computed which characterizes all the pixels that belong to the background, which includes approximately 98\% of the voxels.  The background noise model is computed independently for each slice to account for variations in tissue and imaging properties.  The background model $p_B$ is assumed to be a Gaussian distribution, whose mean and variance $(\mu_B,\sigma_B^2)$ are empirically computed from each slice $z$ (the $z$ index is omitted in Eq 1 for simplicity of notation).  Then, the probability of a voxel belonging to the background, i.e. not being `bright,' is given by
\begin{equation} 
p_B(x,y,z) = \frac{1}{\sigma_B \sqrt{2 \pi}} \int^{\infty}_{v(x,y,z)} e^{\frac{-(t - \mu_B)^2}{2 \sigma_B^2}} d t.
\end{equation} 
Therefore, the probability of a voxel associated with the foreground, $p_F$, is computed as
\begin{equation}
p_F(x,y,z) = 1 - p_B(x,y,z).
\end{equation}
Fig \ref{fig:raw_prob} shows an example of the transformation from the raw data to the foreground probability map. 

\subsection*{\textit{Step 2: Probability of 2D Puncta}}
\begin{figure}[!h]
\includegraphics[width=1\textwidth]{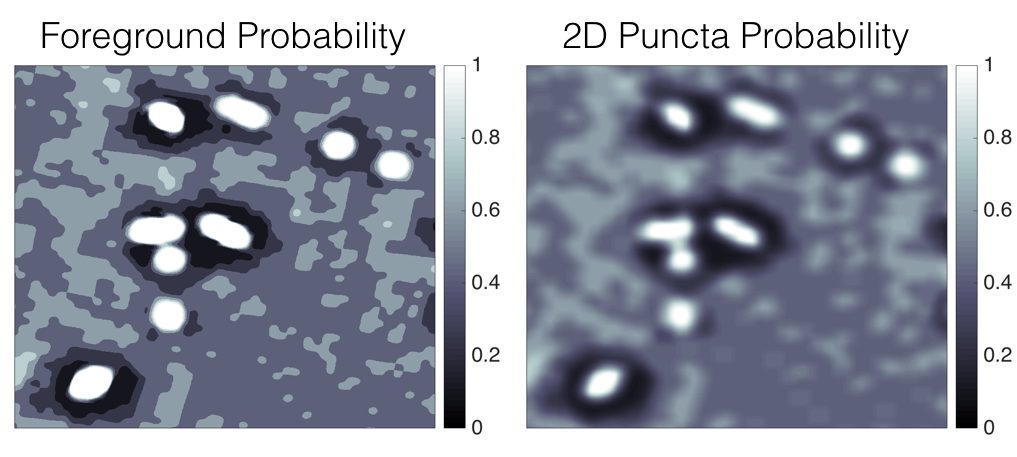}
\caption{\textbf{2D Puncta Probability} - The image on the left is the output from step one, a portion of a single slice of PSD-95 where each pixel codes the probability that it represents signal, not noise.  The image on the right is the result of the corresponding probability map of each pixel belonging to a 2D punctum.  Both images are cutouts of size $2261 \times 2501$ pixels or $5.268 \times 5.827 \mu$ m. }
\label{fig:prob_conv}
\end{figure}

Once foreground pixels have been identified in a probabilistic fashion, the next step is to determine if they form a 2D punctum.  Since synapses appear as bright puncta in the IF image data, voxels which form puncta should have a higher probability of being associated with a synapse than those which do not.  The probability of a voxel belonging to a 2D punctum, $p_P$, is computed by multiplying the voxel's foreground probability by that of its neighbors in a predefined neighborhood region,
\begin{equation} 
p_P(x, y, z) = \prod_{i=x-W}^{x+W} \prod_{j=y-W}^{y+W} p_F(i,j, z),
\label{eq:2dpuncta}
\end{equation} 
where $W$ is the neighborhood size, defined by the smallest expected punctum size.  These operations are analogous to applying a box filter on the logarithm of the probability map, for computational efficiency. In our experiments, $W$ was set to be slightly larger than the size of the point spread function (PSF) of the microscopes used.
\newline
\newline
Fig~\ref{fig:prob_conv} shows an example of the foreground probability map and the corresponding 2D puncta probability map. This step transforms each pixel's value from representing the probability it belongs to the foreground to the probability it belongs to both the foreground \textit{and} to a punctum. 

\subsection*{\textit{Step 3: Probability of 3D Puncta}}
\begin{figure}
\includegraphics[width=1\textwidth]{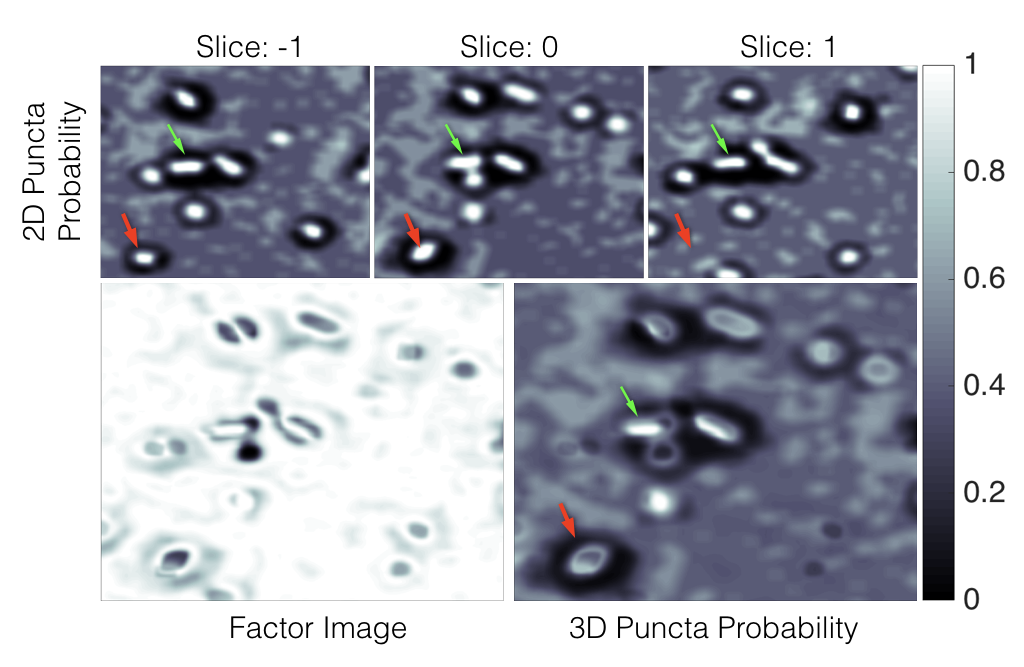}
\caption{\textbf{3D Puncta Probability} - \textit{Top:} Three consecutive slices of 2D puncta probability, given by Eq.~\eqref{eq:2dpuncta} . \textit{Bottom:} Factor image given by Eq.~\eqref{eq:factor} (left) and the corresponding 3D puncta probability, given by Eq.~\eqref{eq:p3dp}, (right) of the center slice in the top row. Only those 2D puncta that actually span multiple slices are kept with high intensity (probability) in the combined result (bottom right).  The green arrow points to an example of a probable punctum that spans multiple slices.  The red arrow points to an example of a relatively-less probable punctum which does not span multiple slices and therefore is diminished in the output image.  Each image is a cutout of size $2261 \times 2501$ pixels or $5.268 \times 5.827 \mu$ m.}
\label{fig:factor_result}
\end{figure}

Potential synaptic puncta can span multiple slices of a given channel; puncta that span multiple slices have a higher probability of being associated with a synapse than those that do not. Therefore, we propose a factor $f(x,y,z)$ which diminishes the probability values associated with voxels which do not maintain a similar probability value in adjacent slices,

\begin{equation} 
f(x, y, z) = \exp \left \{- \sum_{j=j_{start}}^{j=j_{end}} [p_P(x, y, z) - p_P(x, y, z+j)]^2 \right \}.
\label{eq:factor}
\end{equation} 
The pixel's 2D puncta probability is compared to that of its neighbor in slice(s) before, $j_{start}$, and slice(s) after, $j_{end}$.  The number of slices compared is dependent on the input size parameter for each antibody.  The factor attenuates  values for 2D puncta that do not span the required number of slices, as shown in Figure~\ref{fig:factor_result}.  
\newline
\newline
The 3D puncta probability map is then computed by multiplying the 2D puncta probability map by this factor, 

\begin{equation}
p_{3DP}(x,y,z)= p_P(x,y,z) f(x,y,z),
\label{eq:p3dp} 
\end{equation}
which further improves the probability of a detection by considering the slice-to-slice spatial distribution, going from 2D to 3D.

\subsection*{\textit{Step 4: Adjacency of Presynaptic and Postsynaptic Puncta}}
\begin{figure}
\centering
\includegraphics[width=1\textwidth]{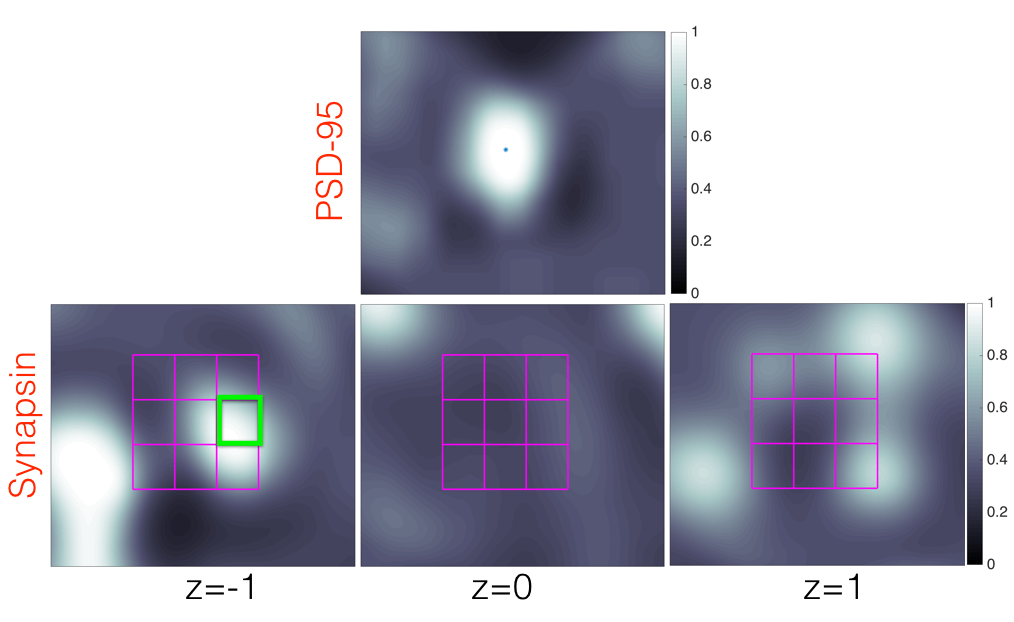}
\caption{
\textbf{Presynaptic and Postsynaptic Puncta Adjacency} - The first row contains a cutout showing a PSD-95 punctum with a pixel highlighted in the center of the image.  The second row contains synapsin cutouts with the search grid overlaid.  For this example, $K=3$, so shown is a $3 \times 3$ grid spanning 3 slices is shown.  The brightest box is highlighted in green.  Thus, the output value of the synaptic probability map at the pixel specific in the PSD-95 image is the average pixel value of the green box multiplied by the intensity value of the PSD-95 pixel.  
}
\label{fig:search_grid}
\end{figure}

\begin{figure}
\centering
\includegraphics[width=0.75\textwidth]{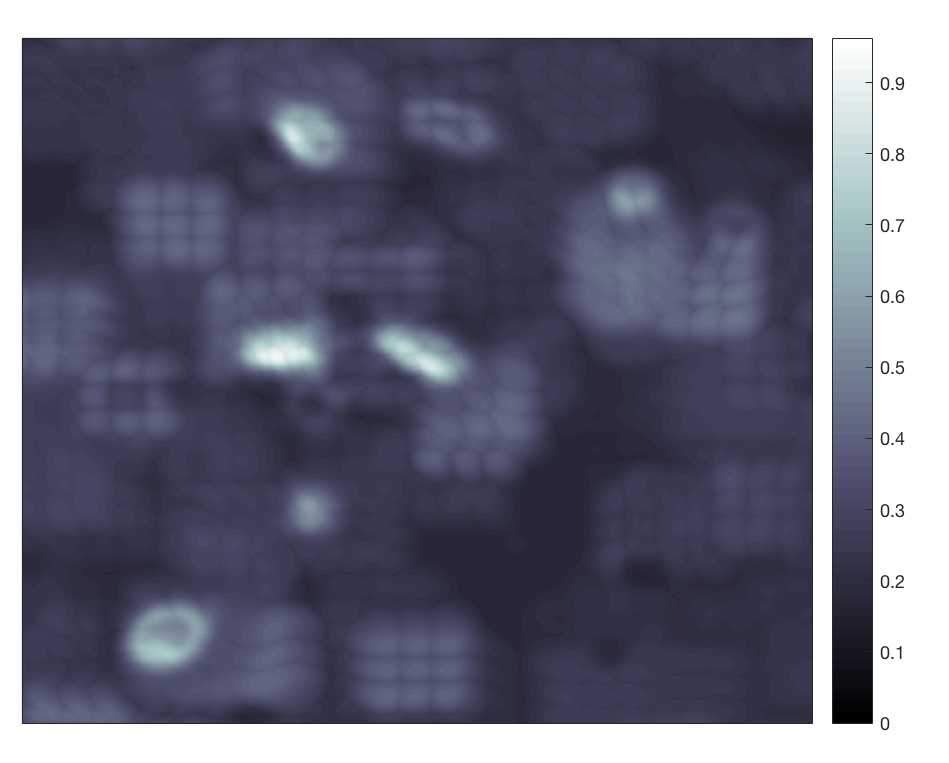}
\caption{
\textbf{Synapse Probability Map} - The output of the method, as described in Eq \ref{eq:finalEq}, where value at each voxel is the probability it belongs to a specific synapse subtype.  Cutout of size $2261 \times 2501$ pixels or $5.268 \times 5.827 \mu$ m.
}
\label{fig:silane_result}
\end{figure}

In electron microscopic images, synapses are identified by the presence of synaptic vesicles on the presynaptic side, the close adjacency of the membranes of the presynaptic axon terminal to a postsynaptic dendrite or dendritic spine, and the presence of a distinct postsynaptic specialization, as diagrammed in Fig \ref{fig:Chemical_Synapse}.  Synapses are identified in IF data by the close spatial arrangement of pre- and postsynaptic antibody markers, which correspond to proteins associated with synapses.  Therefore, the next step in our approach is to look for the presence of presynaptic puncta in the neighborhood of postsynaptic puncta.  More precisely, for each postsynaptic antibody voxel (i.e., PSD-95), we search in the adjacent 3D neighborhood of the corresponding presynaptic (i.e., synapsin) volume for a high intensity probability signal. To accomplish this, a rectangular grid is defined in the presynaptic channels around each postsynaptic voxel, as shown in Fig \ref{fig:search_grid}.  The size of the grid is defined by the initial query parameters, which depend on both the inherent biology and microscope resolution.  The logarithm of the 3D puncta probability map (Eq. \ref{eq:p3dp}) is integrated in each grid location and the maximum is taken as presynaptic signal level around the given postsynaptic location,
\begin{equation} 
\log p_{pres} = \max_{k} \left( \sum_{G_k} \log p_{3DP}(presynaptic) \right),
\end{equation} 
where the grid $G$ is centered at the current voxel $(x,y,z)$ and divided into $K \times K  \times K$ subregions $G_k$.  To search in a grid around a defined voxel location for the presynaptic signal, $K$ is set to 3.  When searching for the postsynaptic signal, $K$ is set to 1 since postsynaptic signals are expected to loosely co-localize. These values can be adopted to the data resolution. 
The postsynaptic antibody pixel probability 
\begin{equation}
p_{post}=p_{3DP}(postsynaptic),
\end{equation}
is multiplied by the presynaptic probability to obtain the desired probability map:

\begin{equation} 
p_{synap}(x,y,z) = p_{pres}(x,y,z) p_{post}(x,y,z).
\label{eq:finalEq} 
\end{equation} 
Again, the probability information is here maintained (Fig \ref{fig:silane_result}), now including the morphological relationship between the channels.  This `grid' like approach allows the method to be robust to slight image alignment and registration issues, as well as to deconvolution artifacts. 

\section*{Experimental Evaluation}

The proposed method was evaluated on a series of array tomography (AT) datasets published in\cite{Collman} and \cite{Weiler}.  These datasets were acquired using the AT methods described in \cite{Micheva2}.   Each dataset was stained and imaged with antibodies for presynaptic and postsynaptic proteins and then aligned and registered.  In the conjugate AT (cAT) dataset \cite{Collman}, the tissue samples were also imaged with a scanning electron microscope (SEM) and then the IF data were up-sampled, aligned, and registered to the EM data.  Synapses identifiable in the EM image data were labeled and used to provide ground truth.  Table \ref{table:antibodyList} lists the synaptic markers used.  

\begin{table}
\centering
\footnotesize
\setlength{\tabcolsep}{8pt}
\centering
\begin{tabular}[h]{c c c}
\toprule
   				&{\textbf{Presynaptic}} 	&{\textbf{Postsynaptic}} \\
\midrule
\textbf{Excitatory} 	& synapsin 	     		&  PSD-95	\\ 
				& VGluT1	 			&  NR1		\\
				& VGluT2	 			&  NR2B		\\
				& 		 			&  GluR1		 \\
				& 	 				&  GluR2		 \\
\cmidrule{1-3}			
\textbf{Inhibitory}	& synapsin 			&  gephyrin	\\ 
 				& GABA	 			&  GABAAR	\\ 	
				& vGAT				& 			\\
				& GAD				&			\\		
\bottomrule
\end{tabular}
\caption{Synaptic markers used in this work across the various datasets.  Not all markers were present in each dataset \cite{Collman} \cite{Weiler}.}
\label{table:antibodyList}
\end{table}

\begin{table}
\centering
\footnotesize
\setlength{\tabcolsep}{8pt}
\centering
\begin{tabular}[h]{c c c c}
\toprule
   						&{\textbf{Dimension (Pixels)}} 		&{\textbf{Resolution}} 			&{\textbf{Labeled Synapses}} \\
\midrule
\textbf{KDM-SYN-120905} 	& $4508 \times 6306 \times 27$ 	&  $2.33 \times 2.33 \times 70$ nm/pixels 	& 236 \\ 
\cmidrule{1-4}			
\textbf{KDM-SYN-140115}		& $7936 \times 9888 \times 39$  	&  $3.72 \times 3.72 \times 70$ nm/pixels	& 1457\\ 
\bottomrule
\end{tabular}
\caption{The cAT datasets used for analysis \cite{Collman}. }
\label{table:cATDatasets}
\end{table}

\subsection*{\textit{Evaluation on Conjugate Array Tomography }}
\begin{figure}[!h]
\centering
\includegraphics[width=1\textwidth]{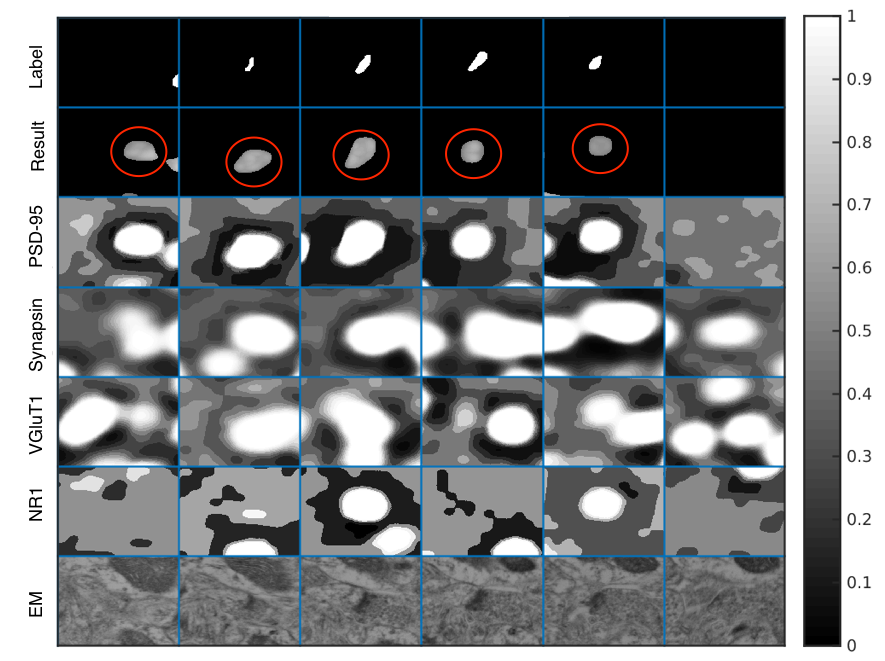}
\caption{\textbf{True Positive Synaptogram} - Synaptogram showing the distribution of IF data for an EM identified synapse.  Each column represents a single slice of data in the $z$ direction, for a total of six slices.  The first row (from the top) shows a manually labeled synaptic cleft, as identified in the EM volume.  The EM data was used only for validation, since the method operates solely on the IF data.  The second row shows the thresholded output of the proposed method, circled in red.  Rows 3-6 show the corresponding foreground probability maps for each channel.  PSD-95 is Postsynaptic Density 95, VGluT1 is Vesicular Glutamate Transporter 1, and NR1 is N-methyl-D-aspartate Receptor 1.  PSD-95 and NR1 are both postsynaptic markers and tend to co-localize, while  synapsin and VGluT1 are both presynaptic markers and tend to co-localize.  Each `block' is $1.221 \mu m \times 1.233 \mu m$. }
\label{fig:synaptogramSilane}
\end{figure}

\begin{figure}[!h]
\centering
\includegraphics[width=1\textwidth]{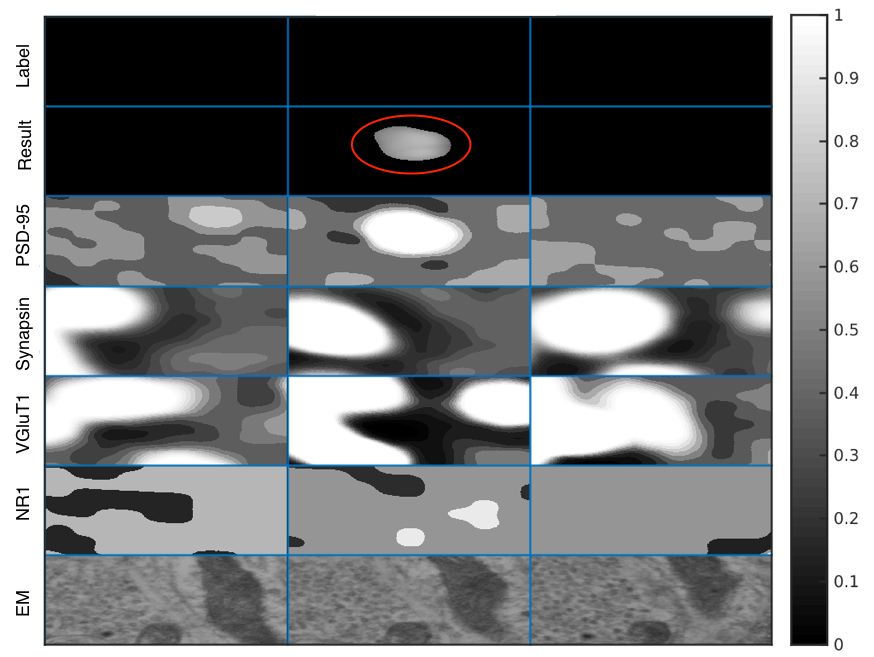}
\caption{\textbf{False Positive Synaptogram} - Synaptogram showing a `false positive.'  Presynaptic and postsynaptic proteins are visible and experts would often label this a synapse presented with IF data alone, but no synapse was identified in the corresponding EM  section. The algorithm makes the same mistake as a human expert.  Each `block' is $1.069 \mu$ m $\times 1.011 \mu$ m.}
\label{fig:synaptogramFP}
\end{figure}

\begin{figure}[!h]
\centering
\includegraphics[width=1\textwidth]{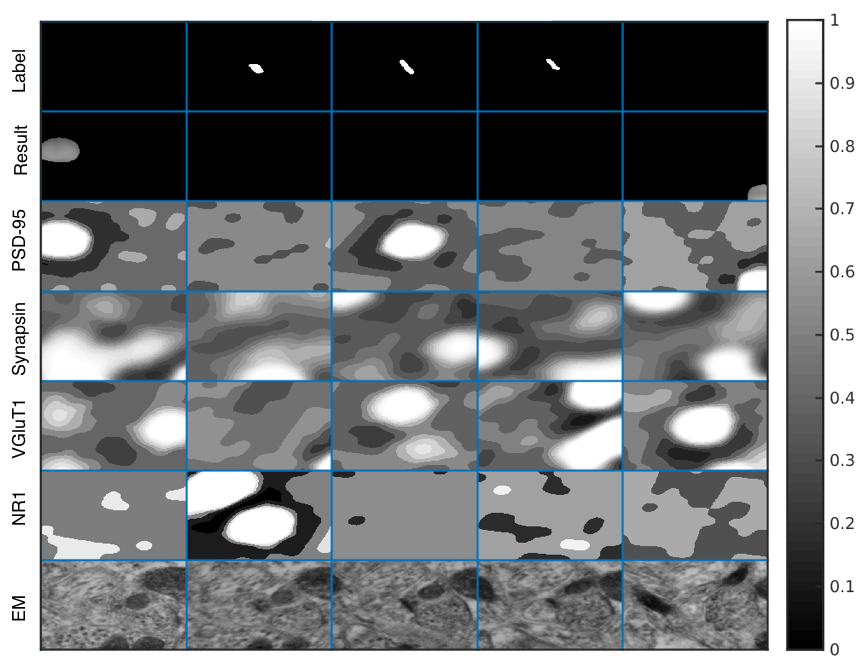}
\caption{\textbf{False Negative Synaptogram} - Synaptogram showing a `false negative.'  While the corresponding EM sections shows a synapse, there is a lack of synaptic IF signal available to justify the presence of a synapse using solely IF data. Again, the algorithm makes the same mistake a human expert would make when working only with the IF data.  Each `block' is $1.086 \mu$ m $\times 1.130 \mu$ m. }
\label{fig:synaptogramFN}
\end{figure}

{\bf Experimental Setup} \\ 
The proposed method was first evaluated on the cAT dataset published in \cite{Collman} using the associated EM image data to create the `ground truth' needed for evaluation.  The datasets themselves are described in Table \ref{table:cATDatasets}.   To evaluate the method's performance on excitatory synapses, the set of query parameters in  Table~\ref{table:excitatoryQueries} were used.  For inhibitory synapse detection, the queries listed in Table~\ref{table:inhibitoryQueries} were used.  These parameters were based on prior literature concerning synaptic proteins and their respective antibodies \cite{Busse} \cite{Weiler}.  Only 20 inhibitory synapses were manually identified in the KDM-SYN-120905 dataset; therefore, inhibitory synapse detection performance is only reported for the larger KDM-SYN-140115 dataset.  For evaluation and visualization purposes, the output probability map, $p_{synap}(x,y,z)$, from each query was thresholded and adjacent voxels that lie over the threshold were grouped into detections.  
\begin{table}
\centering
\footnotesize
\setlength{\tabcolsep}{8pt}
\centering
\begin{tabular}[h]{c c c c c}
\toprule
   & \multicolumn{2}{c}{\textbf{Presynaptic}} & \multicolumn{2}{c}{\textbf{Postsynaptic}} \\
\cmidrule{2-5}
\textbf{Query}   & \textbf{Antibody} & \textbf{Puncta Size} (x,y,z) $\mu m$ & \textbf{Antibody} & \textbf{Puncta Size} (x,y,z) $\mu$ m  \\
\midrule
1 	& synapsin 	     	& 0.2 x 0.2 x 0.21 		& PSD-95 		& 0.2 x 0.2 x 0.21 			\\ 
\cmidrule{1-5}
2	& synapsin 		& 0.2 x 0.2 x 0.21 		& PSD-95			& 0.2 x 0.2 x 0.07			\\ 
 	& VGluT1	 		& 0.2 x 0.2 x 0.21 		& 		 		& 						\\ 
\cmidrule{1-5}
3	& synapsin 		& 0.2 x 0.2 x 0.21 		& PSD-95			& 0.2 x 0.2 x 0.07			\\ 
 	&   		 		&  			 		& NR1	 		& 0.2 x 0.2 x 0.07			\\ 
\cmidrule{1-5}
4	& synapsin 		& 0.2 x 0.2 x 0.21 		& PSD-95			& 0.2 x 0.2 x 0.07			\\ 
 	& VGluT1	 		& 0.2 x 0.2 x 0.21 		& NR1	 		& 0.2 x 0.2 x 0.07			\\ 
\cmidrule{1-5}		
5	& synapsin 		& 0.2 x 0.2 x 0.21 		& PSD-95			& 0.2 x 0.2 x 0.07			\\ 
 	& VGluT1	 		& 0.2 x 0.2 x 0.07 		&  		 		& 	 					\\ 
\cmidrule{1-5}		
6	& synapsin 		& 0.2 x 0.2 x 0.07 		& PSD-95			& 0.2 x 0.2 x 0.07			\\ 
 	& VGluT1	 		& 0.2 x 0.2 x 0.21 		&  		 		& 						\\ 
\cmidrule{1-5}		
7	& synapsin 		& 0.2 x 0.2 x 0.21 		& PSD-95			& 0.2 x 0.2 x 0.07			\\ 
 	& 				& 					& NR1	 		& 0.2 x 0.2 x 0.21			\\ 
\bottomrule
\end{tabular}
\caption{Excitatory synapse detection queries for the cAT data.  Note that the size dimension in $x, y$ correspond to the window width $W$ in Eq. ~\eqref{eq:2dpuncta} and the $z$ range corresponds to the number of slices, $j$, mentioned in Eq.~\eqref{eq:factor}.}
\label{table:excitatoryQueries}
\end{table}
\begin{table}
\centering
\footnotesize
\setlength{\tabcolsep}{8pt}
\centering
\begin{tabular}[h]{c c c c c}
\toprule
   & \multicolumn{2}{c}{\textbf{Presynaptic}} & \multicolumn{2}{c}{\textbf{Postsynaptic}} \\
\cmidrule{2-5}
\textbf{Query}   & \textbf{Antibody} & \textbf{Puncta Size} (x,y,z) $\mu m$ & \textbf{Antibody} & \textbf{Puncta Size} (x,y,z) $\mu$ m  \\
\midrule
1 	& synapsin 	     	& 0.2 x 0.2 x 0.21 		& gephyrin 		& 0.2 x 0.2 x 0.07 			\\ 
	& GABA	 		& 0.2 x 0.2 x 0.07 		& 		 		& 						\\
\cmidrule{1-5}			
2	& synapsin 		& 0.2 x 0.2 x 0.21 		& gephyrin		& 0.2 x 0.2 x 0.07			\\ 
 	& VGAT	 		& 0.2 x 0.2 x 0.07 		& 		 		& 						\\ 
\cmidrule{1-5}			
3	& synapsin 		& 0.2 x 0.2 x 0.21 		& gephyrin		& 0.2 x 0.2 x 0.07			\\ 
 	& GAD  	 		& 0.2 x 0.2 x 0.07  		& 	 			&					 	\\ 		
\bottomrule
\end{tabular}
\caption{Inhibitory synapse detection queries for the cAT data.}
\label{table:inhibitoryQueries}
\end{table}
\newline
\newline
{\bf Performance Metrics } \\
The ground truth is created with EM data; therefore, some of the synapses do not appear in the IF data. Likewise, there are examples of false detections that experts cannot distinguish from true synapses without using the EM data. The proposed approach is unable to handle such cases correctly.
\newline
\newline
We report in Table \ref{table:results} the precision and recall values obtained for these two tested datasets. We differentiate two cases: first, considering all synapses manually identified in the EM data and counting all false positives returned by the program (referred to in Table \ref{table:results} as \textit{`EM'}); and second, considering the subsection of detections that can be manually verified by an expert using only IF data (referred to in Table \ref{table:results} as \textit{`IF'}).  For example, detections that the EM data lists as a false positives but are impossible to verify using only IF data are removed from evaluation.  Similarly, manually-identified synapses in the EM data which do not appear in IF data are also removed from secondary evaluation.  
\begin{table}
\centering
\footnotesize
\setlength{\tabcolsep}{4pt}
\centering
\begin{tabular}[h]{c c c c c | c c c c}
\toprule
\cmidrule{2-9}
& \multicolumn{4}{c}{\textbf{Excitatory}} & \multicolumn{4}{c}{\textbf{Inhibitory}} \\
\cmidrule{2-9}
& \multicolumn{2}{c}{\textbf{EM}} & \multicolumn{2}{c}{\textbf{IF}} & \multicolumn{2}{c}{\textbf{EM}} & \multicolumn{2}{c}{\textbf{IF}}  \\
\cmidrule{2-9}
{\bf Dataset} 	& {\bf Precision} 	& {\bf Recall}  & {\bf Precision} 	& {\bf Recall} & {\bf Precision} 	& {\bf Recall}  & {\bf Precision} 	& {\bf Recall} \\ 
\midrule
KDM-SYN-120905 	& 0.88 	& 0.91 	& 0.90 	& 0.93 	& - 		& - 		& - 		& - \\
KDM-SYN-140115 	& 0.92 	& 0.94 	& 0.93 	& 0.95 	& 0.82 	& 0.81 	& 0.91 	& 0.91 \\
\bottomrule
\end{tabular}
\caption{Excitatory and inhibitory synapse detection results.  Precision is defined as the number of true positives detections / (true positive detections + false positive detections).  Recall is defined as the number of true synapses detected / (true synapses detected + missed synapses). }
\label{table:results}
\end{table}
\begin{table}
\centering
\footnotesize
\setlength{\tabcolsep}{4pt}
\centering
\begin{tabular}[h]{c c c c}
\toprule
 \multicolumn{3}{c}{\textbf{Excitatory, EM}}  \\
{\bf Dataset} 	& {\bf Precision} 	& {\bf Recall}   \\ 
\midrule
KDM-SYN-120905 	& 0.96	& 0.88  \\
KDM-SYN-140115 	& 0.94 	& 0.82   \\
\bottomrule
\end{tabular}
\caption{State-of-the-art detection results for excitatory synapses from \cite{Collman}.}
\label{table:SOAresults}
\end{table}
\begin{figure}[!h]
\centering
\includegraphics[width=1\textwidth]{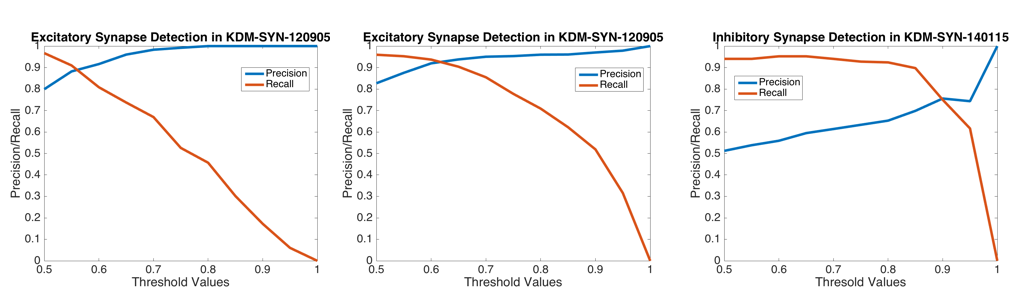}
\caption{\textbf{Precision/Recall Curves} - The relationship between the precision and recall values across a series of thresholds for each cAT dataset.   As detailed in the text, the threshold is for validation purposes, since the proposed framework outputs a confidence/probability. }
\label{fig:pr_curves}
\end{figure}
\newline
\newline
{\bf Results } \\
 Once the final probability map for each query was computed, maps for excitatory synapses were thresholded at 0.6 for the KDM-SYN-140115 dataset and 0.55 for the KDM-SYN-120905 dataset.  The maps for inhibitory synapses were thresholded at 0.9.  These thresholds were based on the intersection of the precision/recall curves in Figure \ref{fig:pr_curves}.  The discrepancy between threshold values is attributed to the different signal/noise distributions of each antibody.  Note that this threshold, the only non-biological parameter of the system, can be ignored when working directly on the output (Eq. \ref{eq:finalEq}), or easily set for the entire dataset by visually inspecting a few detections.  Fig \ref{fig:pr_curves} shows the relationship between the final threshold and accuracy in greater detail.  As shown in Table \ref{table:results}, the proposed algorithm successfully detects most synapses in both datasets, with only a small fraction of false positive detections. Based on the \emph{IF only} indicator, we observe that the algorithm performs at human level (approximately 90\% accuracy), with false positives and false negatives limited to cases which human experts (including co-authors of this manuscript) are also not confident of their own result \cite{Collman}. 
\newline
\newline
As shown in Table~\ref{table:SOAresults}, the proposed algorithm performs as well as the state-of-the-art method \cite{Collman} for excitatory synapse detection, while eliminating the need to undergo the labor-intensive process of cultivating a training dataset.  Furthermore, due to the approximate ten-to-one ratio of excitatory to inhibitory synapses, creating training sets for inhibitory synapses is difficult.  Our method is insensitive to the number of synapses per class as it only returns possible synapses which match the query parameters.  Finally, the fact that we can skip training also makes the proposed system more applicable to diverse datasets without the need for re-design the entire process, as here demonstrated.
\newline
\newline
Fig ~\ref{fig:synaptogramSilane} shows an example of a true positive detection of excitatory synapses in the KDM-SYN-120905 dataset.  The figure shows an example of a `synaptogram,' where each row (third to sixth rows) shows a different channel of immunofluorescent signal and each column is a 2D slice. The first row marked as \emph{Label} shows the manual annotation of the synaptic cleft, i.e., the ground-truth, and the second row, marked as \emph{Result}, corresponds to the output of the proposed synapse detection algorithm.  Rows 3-6 are corresponding sections of each channel's foreground probability map (the output of Step 2).  The seventh row, marked as \emph{EM}, shows the corresponding EM data. Figure~\ref{fig:synaptogramFP} shows an example of a false positive which cannot be differentiated from a real detection by an expert without the assistance of EM data (not available for our algorithm). Figure~\ref{fig:synaptogramFN} shows a similar situation for a false negative detection.

\subsection*{\textit{Evaluation on Array Tomography}}
\begin{figure}[!h]
\includegraphics[width=1\textwidth]{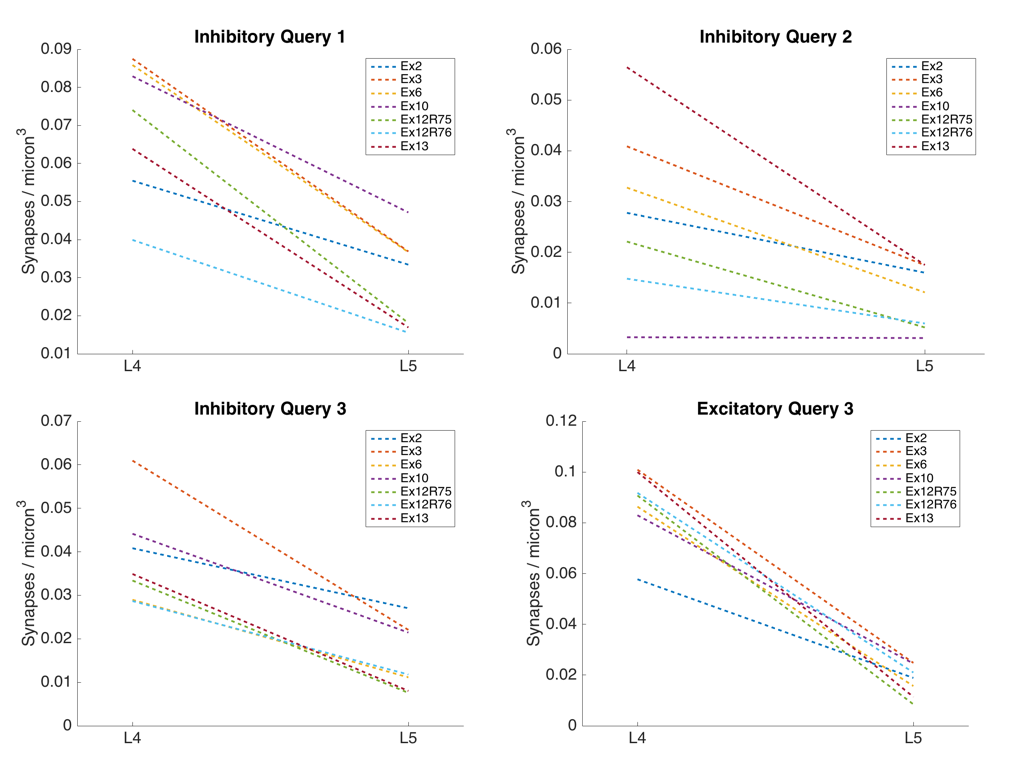}
\caption{\textbf{Mouse Barrel Cortex Layer Density Differences} - The difference in synaptic density between Layer IV and Layer V for specific queries.}
\label{fig:layerComparison}
\end{figure}

The proposed method was evaluated on the array tomography dataset published in \cite{Weiler}, which contains a portion of the mouse barrel cortex from Layer 3 to Layer 5.  Unlike the conjugate array tomography dataset, there is no associated EM imagery. This larger series of datasets includes 11 volumes representing a total of 2,306,233 $\mu m^3$  of cortical volume. Since no gold standard is available for these data, the proposed method was evaluated by verifying known properties of the dataset: there is an approximately ten-to-one ratio of excitatory to inhibitory synapses \cite{Knott_2002}, and there are more inhibitory synapses in Layer 4 than Layer 5 in the mouse barrel cortex \cite{DeFelipe}.  
\newline
\newline
For this dataset, the query parameters were adjusted to reflect the different synaptic markers used.   Tables \ref{table:CB_Excitatory} and \ref{table:CB_Inhibitory} list the query parameters used for detecting both inhibitory and excitatory synapses, similar to those in \cite{Busse} \cite{Collman} \cite{Weiler}.  
\begin{table}
\centering
\footnotesize
\setlength{\tabcolsep}{8pt}
\centering
\begin{tabular}[h]{c c c c c}
\toprule
   & \multicolumn{2}{c}{\textbf{Presynaptic}} & \multicolumn{2}{c}{\textbf{Postsynaptic}} \\
\cmidrule{2-5}
\textbf{Query}   & \textbf{Antibody} & \textbf{Puncta Size} (x,y,z) $\mu m$ & \textbf{Antibody} & \textbf{Puncta Size} (x,y,z) $\mu$ m  \\
\midrule
Query 1 	& synapsin 	     	& 0.2 x 0.2 x 0.14 		& PSD-95 		& 0.2 x 0.2 x 0.14 			\\ 
\cmidrule{1-5}			
Query 2	& synapsin 		& 0.2 x 0.2 x 0.14 		& PSD-95			& 0.2 x 0.2 x 0.07			\\ 
 		& VGluT1	 		& 0.2 x 0.2 x 0.07 		& 		 		& 						\\ 
\cmidrule{1-5}			
Query 3	& synapsin 		& 0.2 x 0.2 x 0.14 		& PSD-95			& 0.2 x 0.2 x 0.07			\\ 
 		& VGluT2  		& 0.2 x 0.2 x 0.07   		& 		 		& 						\\ 
\cmidrule{1-5}			
Query 4	& synapsin 		& 0.2 x 0.2 x 0.14 		& PSD-95			& 0.2 x 0.2 x 0.07			\\ 
 		& GluR1	 		& 0.2 x 0.2 x 0.14 		& 		 		& 						\\ 
\cmidrule{1-5}			
Query 5	& synapsin 		& 0.2 x 0.2 x 0.14 		& PSD-95			& 0.2 x 0.2 x 0.07			\\ 
 		& GluR2	 		& 0.2 x 0.2 x 0.07 		&  		 		& 	 					\\ 
\cmidrule{1-5}			
Query 6	& synapsin 		& 0.2 x 0.2 x 0.07 		& PSD-95			& 0.2 x 0.2 x 0.07			\\ 
 		& 		 		& 			 		& NR2B 		 	& 0.2 x 0.2 x 0.07			\\ 
\bottomrule
\end{tabular}
\caption{Excitatory synapse detection queries for the AT data.}
\label{table:CB_Excitatory}
\end{table}
\begin{table}
\centering
\footnotesize
\setlength{\tabcolsep}{8pt}
\centering
\begin{tabular}[h]{c c c c c}
\toprule
   & \multicolumn{2}{c}{\textbf{Presynaptic}} & \multicolumn{2}{c}{\textbf{Postsynaptic}} \\
\cmidrule{2-5}
\textbf{Query}   & \textbf{Antibody} & \textbf{Puncta Size} (x,y,z) $\mu m$ & \textbf{Antibody} & \textbf{Puncta Size} (x,y,z) $\mu$ m  \\
\midrule
Query 1 	& synapsin 	     	& 0.2 x 0.2 x 0.07 		& gephyrin 		& 0.2 x 0.2 x 0.07 			\\ 
		& 		 		& 					& GABAAR		& 0.2 x 0.2 x 0.07  			\\ 
\cmidrule{1-5}			
Query 2	& synapsin 		& 0.2 x 0.2 x 0.07 		& gephyrin		& 0.2 x 0.2 x 0.07			\\ 
 		& vGAT	 		& 0.2 x 0.2 x 0.07 		& 		 		& 						\\ 
\cmidrule{1-5}			
Query 3	& synapsin 		& 0.2 x 0.2 x 0.07 		& gephyrin		& 0.2 x 0.2 x 0.07			\\ 
 		& GAD  	 		& 0.2 x 0.2 x 0.07  		& 	 			&					 	\\ 
		
\bottomrule
\end{tabular}
\caption{Inhibitory synapse detection queries for the AT data.}
\label{table:CB_Inhibitory}
\end{table}
\begin{figure}[!h]
\centering
\includegraphics[width=1\textwidth]{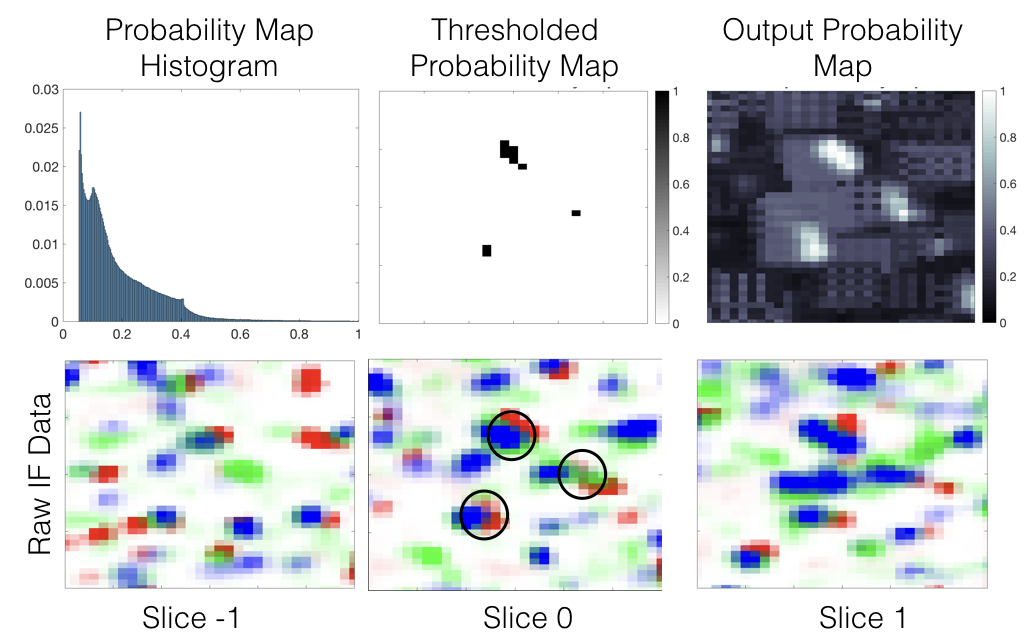}
\caption{\textbf{Synapse Detection Results} - {\it Top Left:} Histogram of a slice of the probability map. {\it Top Middle:} Thresholded probability map, see text for details on the threshold selection. {\it Top right:}  Probability map of one slice of a  $13$ $\mu$ m $\times 17$ $\mu$ m region of the AT data on \cite{Weiler}.  {\it Bottom Row:} . A series of pseudo-color images indicating the presence of the receptors PSD-95 (red), vGluT1 (blue), and synapsin (green) across three consecutive slices.  Centroids of detected synapses containing all three receptors (query) are circled in black. } 
\label{fig:cb_output}
\end{figure}
\begin{figure}[!h]
\centering
\includegraphics[width=1\textwidth]{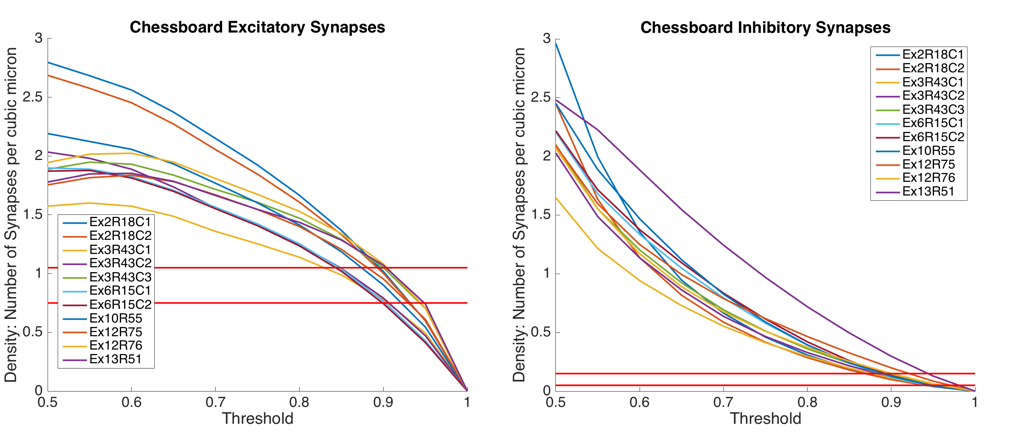}
\caption{\textbf{Thresholds vs Density} - These plots show the variation of punitive synapse density across different thresholds.  Each curve represents a dataset in \cite{Weiler} and the red lines show the expected synaptic density.  For excitatory synapses, the expected density is 0.9 synapses $\mu$ m$^3 \pm 0.15 \mu$ m$^3$.  For inhibitory synapses, the expected density is 0.1 synapses $\mu$ m$^3 \pm 0.05 \mu$ m$^3$ \cite{Calverley} \cite{Schuz}. }
\label{fig:cb_thresholds}
\end{figure}
\newline
\newline
{\bf Thresholding the Probability Map} \\ 
Once the probability maps were computed (Figure \ref{fig:cb_output}), they were thresholded for evaluation purposes only.  Thresholds for each dataset were determined by examining the synaptic density values across various thresholds, as shown in Fig \ref{fig:cb_thresholds}.  As the figure shows, the appropriate thresholds for each dataset exist in a narrow band, consistent with the results in the cAT dataset.  Thresholding at the optimal value shown in Fig \ref{fig:cb_thresholds} for each dataset, as set by plots in Fig \ref{fig:cb_thresholds}, amounted to 2,326,692 excitatory synapses and 252,833 inhibitory synapses.  This amounts to approximately $1.12$ synapses per cubic micrometer and an overall ratio of $9.2$ excitatory to inhibitory synapses, which is consistent with results in the literature \cite{Beaulieu} \cite{Schuz}.  
\newline
\newline
Previous quantitative electron microscopy indicates that the synapse density should be higher in Layer IV than in Layer V \cite{Micheva3}, consistent with the results from our algorithm, as shown in the graphs in Fig \ref{fig:layerComparison}.  For all three inhibitory synaptic queries, there is a synapse density difference of more than 50\% between Layer IV and Layer V.   There is also a greater than 50\% synaptic density difference between Layer IV and Layer V for excitatory synapses containing vGluT2, as supported by \cite{Busse} \cite{Graziano} \cite{Nakamura}.   These results further support the validity of the proposed method by confirming known biological properties of a large dataset.  
\newline
\newline
The threshold of the estimated probability can be set to optimize a specific desired property (density in this case), thereby becoming an additional `query.' The threshold can actually add flexibility, since different thresholds might lead to selective detection of different types of synapses. This possibility will be studied when new data becomes available, now that the unsupervised algorithm here introduced can be applied to such data (previous algorithms were basically limited to making binary decisions for detecting synapses they have been trained to detect). This means that the only non-straightforwardly physical parameter of the proposed algorithm (virtually all image processing algorithms have critical parameters) can add flexibility to the technique. Finally, the threshold can be ignored if we work directly with the probability map (Eq. \ref{eq:finalEq}), e.g., to compute `fuzzy volumes.' This unique aspect of the proposed algorithm output will also be the subject of study when running the algorithm on new AT data in the future.

\section*{Discussion and Conclusion}

This work introduced a model-based unsupervised synapse detection algorithm that incorporates both fundamental biological knowledge of synapses and how they are identified by experts in the immunofluorescence data. For this purpose, we created a series of probabilistic excitatory detectors for various subtypes of synapses, and included the 3D spatial relationships typical of synaptic structures.  This novel approach provides a probabilistic-based detection algorithm yielding not only detection but detection with confidence values. The implementation of synapse detection as a probability map (i.e., probability of each pixel belonging to a synapse), as opposed to a binary detection / no-detection result may provide a powerful tool to assist experts throughout the exploratory process to gain new insights from the immunofluorescence data, including potentially discovering new subtypes of synapses.  The influence of different biological and AT components on the actual probability values, from the noise of the system to the expression level of the proteins and the subclass of the synapses, an important new topic of investigation, will become possible when the proposed algorithm is applied to large new datasets, currently being generated.
\newline
\newline
The algorithm is computationally very simple and the only parameters are the user's definition of a synapse subtype, rendering it ready for massive datasets.  The results were validated with the best available cAT and AT data, producing state-of-the-art results without the need for supervised training.  As demonstrated here, the proposed framework can be exploited for the explicit detection of synapses or their properties, the latter being critical for the discovery of new subtypes as well as the patterns of distributions of known subtypes. These, and the potential consequences of the approach here proposed to other modalities, are the subjects of our current efforts.

\section*{Acknowledgments}

This work was supported by the National Institutes of Health (NIH-TRA 1R01NS092474), the Allen Institute for Brain Sciences (AIBS), the U.S. Office of Naval Research (ONR), the U.S. Army Research Office (ARO), the National Science Foundation (NSF), and the U.S. National Geospatial Intelligence Agency (NGA).

\nolinenumbers


\newpage
\begin{thebibliography}{10}

\bibitem{Beaulieu}
Beaulieu C, Campistron G, Crevier C. Quantitative aspects of the GABA circuitry in the primary visual cortex of the adult rat. Journal of Comparative Neurology. 1994 Jan 22;339(4):559-72.

\bibitem{Bloss} 
Bloss EB, Cembrowski MS, Karsh B, Colonell J, Fetter RD, Spruston N. Structured dendritic inhibition supports branch-selective integration in CA1 pyramidal cells. Neuron. 2016 Mar 2;89(5):1016-30.

\bibitem{Burette} 
Burette A, Collman F, Micheva KD, Smith SJ, Weinberg RJ. Knowing a synapse when you see one. Frontiers in Neuroanatomy. 2015;9.

\bibitem{Busse}
Busse B, Smith S. Automated analysis of a diverse synapse population. PLoS Comput Biol. 2013 Mar 28;9(3):e1002976.

\bibitem{Calverley}
Calverley RK, Jones DG. A serial-section study of perforated synapses in rat neocortex. Cell and Tissue Research. 1987 Mar 1;247(3):565-72.

\bibitem{Collman} 
Collman F, Buchanan J, Phend KD, Micheva KD, Weinberg RJ, Smith SJ. Mapping synapses by conjugate light-electron array tomography. The Journal of Neuroscience. 2015 Apr 8;35(14):5792-807.

\bibitem{Costa} 
da Costa NM, Hepp K, Martin KA. A systematic random sampling scheme optimized to detect the proportion of rare synapses in the neuropil. Journal of Neuroscience Methods. 2009 May 30;180(1):77-81.

\bibitem{DeFelipe} 
De Felipe J, Marco P, Fairen A, Jones EG. Inhibitory synaptogenesis in mouse somatosensory cortex. Cerebral Cortex. 1997 Oct 1;7(7):619-34.

\bibitem{Fitzsimmons} 
Fitzsimmons J, Kubicki M, Shenton ME. Review of functional and anatomical brain connectivity findings in schizophrenia. Current opinion in psychiatry. 2013 Mar 1;26(2):172-87.

\bibitem{Graziano} 
Graziano A, Liu XB, Murray KD, Jones EG. Vesicular glutamate transporters define two sets of glutamatergic afferents to the somatosensory thalamus and two thalamocortical projections in the mouse. Journal of comparative neurology. 2008 Mar 10;507(2):1258-76.

\bibitem{Harris} 
Harris KM, Weinberg RJ. Ultrastructure of synapses in the mammalian brain. Cold Spring Harbor Perspectives in Biology. 2012 May 1;4(5):a005587.

\bibitem{Kim} 
Rah JC, Feng L, Druckmann S, Lee H, Kim J. From a meso-to micro-scale connectome: array tomography and mGRASP. Frontiers in Neuroanatomy. 2015 Jun 4;9:78.

\bibitem{Knott_2002} 
Knott, Graham W., et al. `Formation of dendritic spines with GABAergic synapses induced by whisker stimulation in adult mice.' Neuron 34.2 (2002): 265-273.

\bibitem{Knott_2006} 
Knott GW, Holtmaat A, Wilbrecht L, Welker E, Svoboda K. Spine growth precedes synapse formation in the adult neocortex in vivo. Nature neuroscience. 2006 Sep 1;9(9):1117-24.

\bibitem{Korogod} 
Korogod N, Petersen CC, Knott GW. Ultrastructural analysis of adult mouse neocortex comparing aldehyde perfusion with cryo fixation. Elife. 2015 Aug 11;4.

\bibitem{Lichtman}
Lichtman JW, Denk W. The big and the small: challenges of imaging the brain's circuits. Science. 2011 Nov 4;334(6056):618-23.

\bibitem{Micheva3} 
Micheva KD, Beaulieu C. An anatomical substrate for experience-dependent plasticity of the rat barrel field cortex. Proceedings of the National Academy of Sciences. 1995 Dec 5;92(25):11834-8.

\bibitem{Micheva4} 
Micheva KD, Beaulieu C. Development and plasticity of the inhibitory neocortical circuitry with an emphasis on the rodent barrel field cortex: a review. Canadian Journal of Physiology and Pharmacology. 1997 May 1;75(5):470-8.

\bibitem{Micheva2} 
Micheva KD, Smith SJ. Array tomography: a new tool for imaging the molecular architecture and ultrastructure of neural circuits. Neuron. 2007 Jul 5;55(1):25-36.

\bibitem{Micheva1} 
Micheva KD, Busse B, Weiler NC, O'Rourke N, Smith SJ. Single-synapse analysis of a diverse synapse population: proteomic imaging methods and markers. Neuron. 2010 Nov 18;68(4):639-53.

\bibitem{Nakamura} 
Nakamura K, Watakabe A, Hioki H, Fujiyama F, Tanaka Y, Yamamori T, Kaneko T. Transiently increased colocalization of vesicular glutamate transporters 1 and 2 at single axon terminals during postnatal development of mouse neocortex: a quantitative analysis with correlation coefficient. European Journal of Neuroscience. 2007 Dec 1;26(11):3054-67.

\bibitem{O'Rourke} 
O'Rourke NA, Weiler NC, Micheva KD, Smith SJ. Deep molecular diversity of mammalian synapses: why it matters and how to measure it. Nature Reviews Neuroscience. 2012 Jun 1;13(6):365-79.

\bibitem{Rah2} 
Rah JC, Bas E, Colonell J, Mishchenko Y, Karsh B, Fetter RD, Myers EW, Chklovskii DB, Svoboda K, Harris TD, Isaac JT. Thalamocortical input onto layer 5 pyramidal neurons measured using quantitative large-scale array tomography. Frontiers in Neural Circuits. 2013 Nov 12;7:177.

\bibitem{Schuz}
Schuz A, Palm G. Density of neurons and synapses in the cerebral cortex of the mouse. Journal of Comparative Neurology. 1989 Aug 22;286(4):442-55.

\bibitem{Wang} 
Wang G, Smith SJ. Sub-diffraction limit localization of proteins in volumetric space using Bayesian restoration of fluorescence images from ultrathin specimens. PLoS Comput Biol. 2012 Aug 30;8(8):e1002671. 
 
\bibitem{Weiler} 
Weiler NC, Collman F, Vogelstein JT, Burns R, Smith SJ. Synaptic molecular imaging in spared and deprived columns of mouse barrel cortex with array tomography. Scientific Data. 2014 Dec 23;1.

\end{thebibliography}
\end{document}